\documentclass[letterpaper, 10 pt, conference]{ieeeconf}  % Comment this line out if you need a4paper

\IEEEoverridecommandlockouts                              % This command is only needed if 
\usepackage{amsmath,amsfonts,amssymb,amsbsy,bm}
\usepackage{algorithm}
\usepackage{algorithmic}
\usepackage{graphicx}
\usepackage{booktabs}
\usepackage{xcolor}
\newcommand{\revised}[1]{\textcolor{black}{#1}}
\usepackage{acronym}

\usepackage{amsthm}
\usepackage{cite}
\usepackage{hyperref}
\usepackage{threeparttable}
\usepackage{multirow}

\usepackage[table]{xcolor}
\definecolor{redlight}{RGB}{255,235,230}     % Very soft peach
\definecolor{redmed}{RGB}{254,201,180}       % Pastel orange-pink
\definecolor{reddark}{RGB}{252,160,130}      % Light salmon
\definecolor{redverydark}{RGB}{244,109,67}   % Muted coral (Darkest, but still high contrast with black text)
\definecolor{greenlight}{RGB}{230,245,225}   % Soft mint green
\definecolor{greenmed}{RGB}{185,230,210}     % Pale seafoam
\definecolor{greendark}{RGB}{130,210,190}    % Mid-tone teal
\definecolor{greenverydark}{RGB}{75,180,160}  % Vivid teal (Darkest shade, but highly readable under black text)

\theoremstyle{definition}

\theoremstyle{remark}
\newtheorem*{remark}{Remark}
\acrodef{MPC}[MPC]{Model Predictive Control}
\acrodef{RLS}[RLS]{Recursive Least Squares}
\acrodef{PE}[PE]{Persistent Excitation}
\acrodef{RBF}[RBF]{Radial Basis Function}
\acrodef{DDR}[DDR]{Differential-Drive Robot}
\acrodef{FWMAV}[FWMAV]{Flapping-Wing Micro Aerial Vehicle}
\acrodef{QP}[QP]{Quadratic Program}
\acrodef{DNN}[DNN]{Deep Neural Network}
\acrodef{EDMD}[EDMD]{Extended Dynamic Mode Decomposition}
\acrodef{HAVOK}[HAVOK]{Hankel Alternative View of Koopman}
\acrodef{CR-RKL}[CR-RKL]{Covariance-Regulated Recursive Koopman Learning}
\acrodef{PWM}[PWM]{Pulse Width Modulation}

\title{\LARGE \bf
Covariance-Regulated Recursive Koopman Learning for Nonlinear Systems with Uncertain Time-Varying Dynamics
}

\author{Weibin Gu$^{1,\dagger}$, Chen Yang$^{1,2}$, Lu Shi$^{1}$, Chao Gao$^{1,3,\dagger}$ % <-this % stops a space
\thanks{$^{1}$Tsinghua University, PR China
        }%
\thanks{$^{2}$China University of Petroleum-Beijing at Karamay, PR China
        }%
\thanks{$^{3}$Xinchen Qihang Inc., PR China
        }%
\thanks{$^{\dagger}$Corresponding authors}
\thanks{This work was supported in part by the China Postdoctoral Science Foundation (Grant No. 2025M781650) and Xinchen Qihang Inc., PR China.}
}

\begin{document}

\maketitle
\thispagestyle{empty}
\pagestyle{empty}

%%%%%%%%%%%%%%%%%%%%%%%%%%%%%%%%%%%%%%%%%%%%%%%%%%%%%%%%%%%%%%%%%%%%%%%%%%%%%%%%
\begin{abstract}
Offline models for autonomous robots often fail under time-varying dynamics outside their training distribution. While Koopman operator theory offers a linear representation of nonlinear dynamics via lifting, its transition to online recursive estimation may suffer numerical vulnerabilities: covariance windup under low excitation when using exponential forgetting, and vanishing gain without forgetting. \revised{To prevent covariance explosion and parameter freezing}, this paper introduces a \revised{unified} Covariance-Regulated Recursive Koopman Learning (CR-RKL) framework \revised{that systematically incorporates and evaluates several covariance-regulated methods. Specifically, we propose a novel error dead-zone covariance skipping strategy, along with adaptations of constant-trace normalization and directional forgetting to the Koopman-specific setting.} Validated on a non-holonomic differential-drive robot with wheel slip and Stribeck friction, as well as on a two-winged, tailless flapping-wing micro aerial vehicle, CR-RKL achieves numerically stable and accurate online modeling \revised{relative to its offline and vanilla recursive counterparts}. When embedded in model predictive control, it maintains reliable tracking performance under uncertain, time-varying dynamics with bounded dictionaries. \revised{However, we also identify a structural limitation: while covariance-regulated methods achieve low off-policy prediction error, pairing them with unbounded dictionaries can lead to control failure.}
\end{abstract}

\section{Introduction}
\label{sec:introduction}

Deploying high-performance autonomous robots in unstructured environments demands control-oriented models capable of capturing underlying dynamics that are frequently poorly understood, highly time-varying, and corrupted by multi-source uncertainties. In these scenarios, offline models—whether derived from first-principles analysis or batch-trained black-box approaches—often fail to generalize when the system operates outside the nominal training distribution. To ensure long-term autonomy within dynamic and unstructured environments, online model adaptation becomes indispensable and is particularly crucial for complex, underactuated, or highly non-stationary robotic systems, such as terrestrial vehicles subject to nonlinear tire-terrain interactions and bio-inspired \acp{FWMAV}. While existing online adaptation schemes leveraging \acp{DNN} (see, e.g.,~\cite{nagabandi2020deep}) can mitigate model mismatch, they are typically computationally heavy, sample-inefficient, and difficult to integrate with classical stability analysis, control synthesis, or real-time optimization frameworks. Consequently, learning computationally tractable models of nonlinear dynamical systems with uncertain, time-varying behaviors remains a fundamental challenge in system identification and control.

To bridge the gap between model expressivity and computational tractability, Koopman operator theory has emerged as a mathematically rigorous paradigm~\cite{koopman1931hamiltonian,brunton2022modern,shi2026koopman}. By lifting nonlinear system states into a higher-dimensional space of scalar-valued observable functions, the underlying dynamics admit an approximately linear representation. In practical applications, finite-dimensional approximations of this linear operator are constructed via data-driven identification methods such as \ac{EDMD}~\cite{williams2015data} and \ac{HAVOK}~\cite{brunton2017chaos}. Constructing linear models in the lifted space allows advanced optimal control strategies such as \ac{MPC} to be formulated as convex optimization problems that can be solved efficiently in real time. 

To maintain model accuracy during deployment, a number of studies have explored online modifications to Koopman identification. Early approaches relied on periodic batch retraining or iterative numerical optimization to update the operator matrices~\cite{abraham2019active}, which proved computationally intensive and unsuited for high-rate control loops. To circumvent this bottleneck, fast online identification methods motivated by \ac{RLS} algorithm have been proposed to update the Koopman operator matrices recursively at each sampling time using the Sherman-Morrison-Woodbury matrix inversion formula, hence only require the current snapshot, the prior estimate, and a stored covariance matrix~\cite{zhang2019online}. Subsequent work has introduced sliding-window Koopman learning~\cite{zhang2019online}, recursive updates executed without a forgetting factor~\cite{zhang2019online,ng2021model,zhangsample}, recursive variants incorporating exponential forgetting factors~\cite{zhang2019online,calderon2021koopman,ozkan2025modeling}, and Kalman-filter-based recursive variant~\cite{liu2026koopman}. Some of these recursive formulations have been successfully validated on robotic systems such as elastic cable suspension, Stewart platform, turtle robot, and off-road vehicle.

\begin{figure}[!t]
    \centering
    \includegraphics[width=0.9\linewidth]{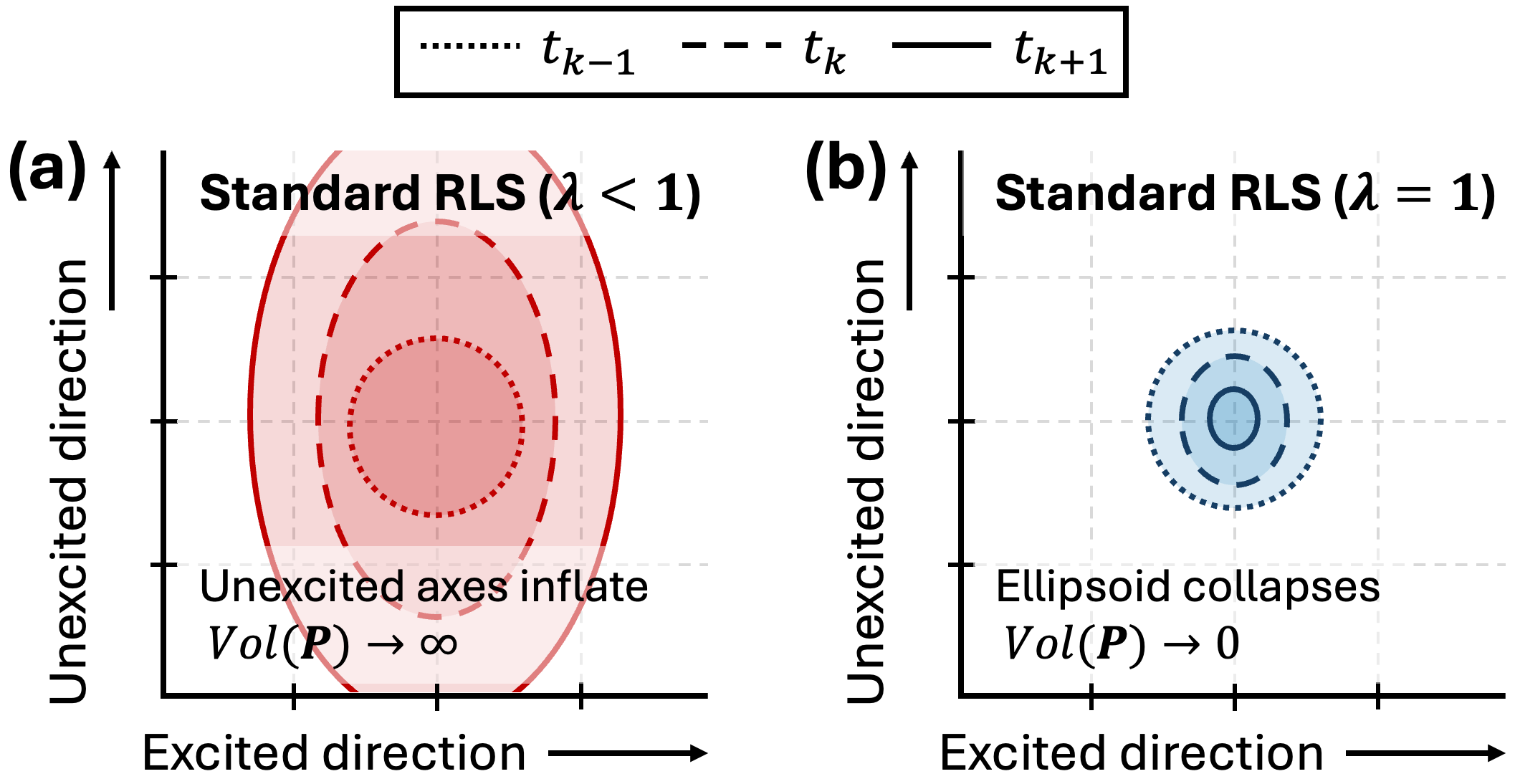}
    \caption{\textbf{Illustration of covariance ellipsoid variations over time under different failure modes in RLS.} \textbf{{(a)}} Standard RLS with an exponential forgetting factor (\(\lambda < 1\)) suffers from covariance windup due to insufficient excitation. \textbf{{(b)}} Standard RLS without a forgetting factor (\(\lambda = 1\)) experiences vanishing gain as covariance matrix drops toward zero.}
    \label{fig:covariance_windup}
    \vspace{-2em}
\end{figure}

Despite these advancements, transitioning Koopman operator identification from offline batch processing to online recursive estimation introduces severe numerical vulnerabilities that are frequently overlooked. Specifically, implementing \ac{RLS} with an exponential forgetting factor introduces a susceptibility to \textit{covariance windup}~\cite{cao2000directional}. When the system executes a low-entropy trajectory or enters a steady-state regime lacking \ac{PE}, the exponential forgetting factor acts as an inflation multiplier on the regularized inverse information matrix, precipitating exponential growth of the estimation error covariance (Fig.~\ref{fig:covariance_windup}(a)). Crucially, this failure mode is intrinsically coupled to the structural properties of the Koopman dictionary. Poorly conditioned or over-parameterized observables introduce severe collinearity, leaving numerous directional manifolds completely unexcited even during dynamic maneuvers. Conversely, while \ac{RLS} without a forgetting factor prevents covariance explosion, it suffers from \textit{vanishing gain}~\cite{landau2011adaptive}, where the adaptation gains monotonically vanish as covariance drops toward zero (Fig.~\ref{fig:covariance_windup}(b)), rendering the model blind to subsequent non-stationary dynamics shifts. Current studies on recursive Koopman learning for robotics rarely address these stability-critical issues systematically, leaving potentially severe failure risks during deployment. A notable exception is found in~\cite{zhangsample}, which establishes a formal convergence analysis of \ac{EDMD} and \ac{RLS} based on Markov chain. However, this theoretical analysis relies on the assumption of an ergodic Markov chain, which may be violated by complex, underactuated robotic systems operating under unconstrained state spaces or unbounded inputs. Moreover, practical regularized solutions to mitigate under-excitation within the recursive Koopman learning is not investigated.

\begin{figure}[!t]
    \centering
    \includegraphics[width=0.9\linewidth]{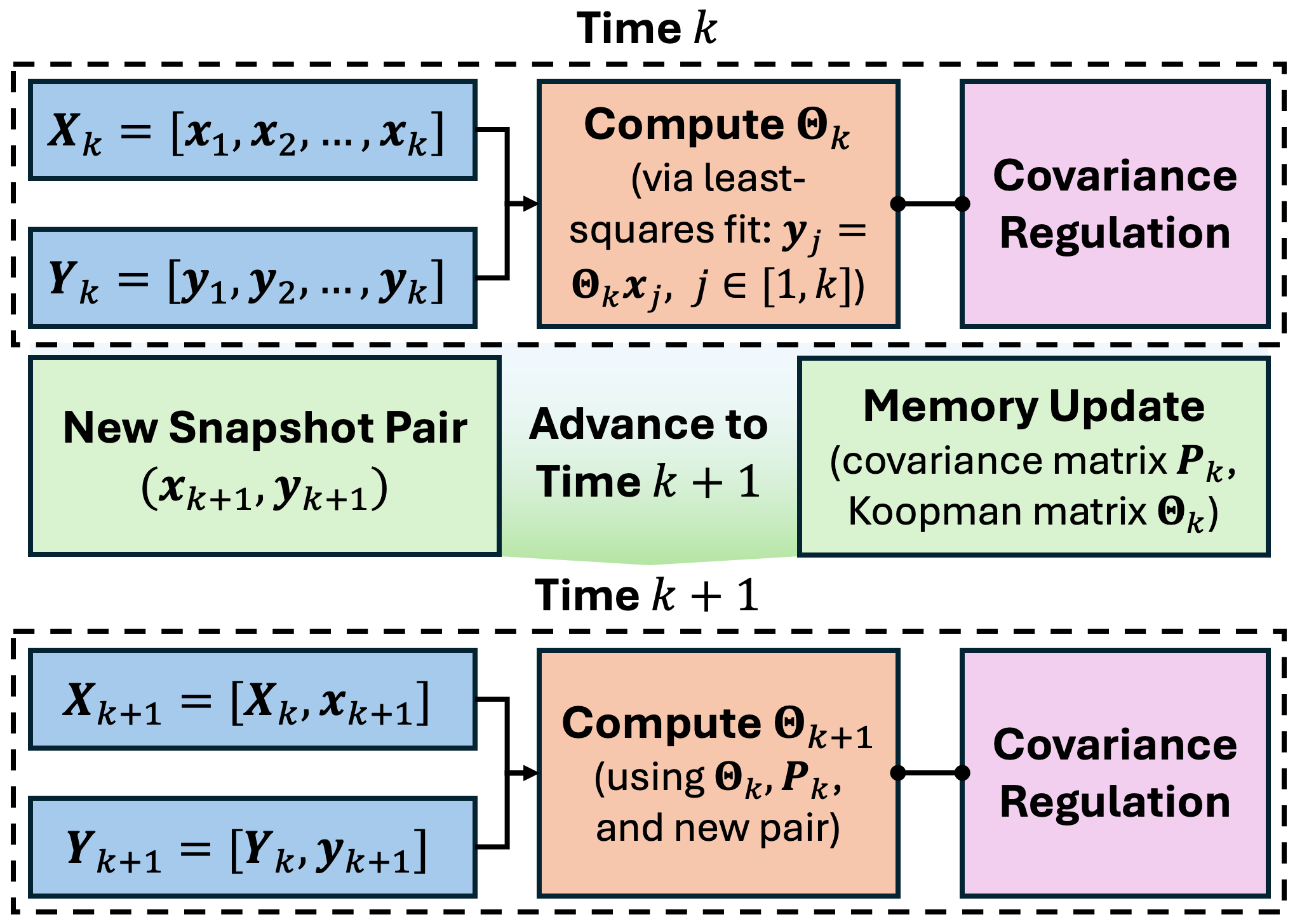}
    \caption{\textbf{The proposed CR-RKL framework.} At each new time step, Koopman operator \(\bm{\Theta}_{k+1}\) is updated in a recursive fashion using the prior operator \(\bm{\Theta}_{k}\) and the stored covariance matrix \(\mathbf{P}_{k}\) from the previous step, alongside the newly observed snapshot pairs \((\mathbf{x}_{k+1}, \mathbf{y}_{k+1})\). $\mathbf{P}_{k+1}$ is regulated via a covariance-regulated strategy (e.g., error dead-zone gating, constant-trace normalization, or subspace-of-information forgetting). Without loss of generality, $\mathbf{X}$ and $\mathbf{Y}$ denote the snapshot matrices at two consecutive time steps.}
    \label{fig:CR-RKL}
    \vspace{-1em}
\end{figure}

To address these coupled numerical vulnerabilities, \revised{this paper introduces a unified \ac{CR-RKL} framework (Fig.~\ref{fig:CR-RKL}) backed by a systematic analysis. We compare several covariance-regulated strategies to address windup and vanishing gain in high-dimensional lifting dictionaries. Specifically, we propose a novel \textit{error dead-zone covariance skipping} strategy, which maintains parameter updates while suspending covariance inflation in contrast to existing error-gating methods that completely skip the Koopman identification~\cite{calderon2021koopman,loya2026koopman}. Furthermore, we adapt established methods—namely \textit{constant-trace normalization}~\cite{airimitoaie2025accelerating} and \textit{subspace-of-information directional forgetting}~\cite{lai2024sift}—into Koopman identification to evaluate their efficacy. While these regulation strategies guarantee numerical survival by bounding the covariance matrix (acting as an adaptation gain matrix rather than a true statistical covariance), they do not inherently guarantee prediction accuracy or closed-loop stability for all dictionaries. In particular, we explicitly highlight a structural limitation: while covariance regulation robustly preserves continuous adaptation for bounded dictionaries, it can cause severe closed-loop tracking failure when paired with unbounded lifting functions.} The proposed \ac{CR-RKL} framework is validated and compared against both offline batch baseline and vanilla online recursive variant across two different robotic platforms (Fig.~\ref{fig:robot}): (i) a simulated non-holonomic \ac{DDR} subject to wheel slip and Stribeck friction dynamics~\cite{pacejka2006tyre,armstrong1990stick}, and (ii) a two-winged, tailless \ac{FWMAV} exhibiting highly nonlinear aerodynamics~\cite{gu202626}. We show consistent results that the \ac{CR-RKL} framework achieves numerically stable and accurate online modeling, and when embedded in model predictive control, it maintains reliable tracking performance for bounded dictionaries under uncertain, time-varying dynamics.

The remainder of this paper is organized as follows. Section~\ref{sec:method} establishes the lifted input-affine Koopman formulation, analyzes the mechanics of main failure modes, and presents three classes of covariance-regulated strategies. Section~\ref{sec:case_study} details the robotic settings under consideration. Section~\ref{sec:results} discusses online Koopman identification and \ac{MPC} tracking results. Lastly, Section~\ref{sec:conclusion} concludes this work.

\section{Methodology}
\label{sec:method}

% We present the \ac{CR-RKL} framework by first formulating recursive update laws for finite-dimensional Koopman operators with arbitrary lifting dictionaries, then analyzes potential failure modes, and introduces three classes of covariance-regulated strategies to guarantee numerical stability.

\subsection{Generalized Recursive Koopman Learning}
Consider a general discrete-time nonlinear robotic system
\begin{equation}
\label{eq:nonlinear_system}
\mathbf{x}_{k+1} = \mathbf{f}(\mathbf{x}_k, \mathbf{u}_k),
\end{equation}
with system state $\mathbf{x}_k \in \mathbb{R}^n$ and control input $\mathbf{u}_k \in \mathbb{R}^m$. The Koopman operator $\mathcal{K}$ advances scalar observables $\psi: \mathbb{R}^n \to \mathbb{R}$ linearly by $\mathcal{K}\psi(\mathbf{x}_k) = \psi(\mathbf{f}(\mathbf{x}_k, \mathbf{u}_k))$. To construct a computationally tractable model for control synthesis, we approximate $\mathcal{K}$ using a dictionary of $N \gg n$ observables $\bm{\Psi}(\mathbf{x}) \in \mathbb{R}^N$ with an input-affine linear model given as
\begin{equation}
\label{eq:lifted_affine}
\bm{\Psi}(\mathbf{x}_{k+1}) \approx \mathbf{A}_k \bm{\Psi}(\mathbf{x}_k) + \mathbf{B}_k \mathbf{u}_k,
\end{equation}
where $\mathbf{A}_k \in \mathbb{R}^{N \times N}$ and $\mathbf{B}_k \in \mathbb{R}^{N \times m}$ denote the state transition and control input matrices, which are time-varying and need to be estimated online.

Now let the unknown linear operators be concatenated into a unified parameter matrix $\bm{\Theta}_k = [\mathbf{A}_k \; \mathbf{B}_k] \in \mathbb{R}^{N \times (N+m)}$, and define the augmented regressor vector as $\bm{\zeta}_k = [\bm{\Psi}(\mathbf{x}_k)^\top, \mathbf{u}_k^\top]^\top \in \mathbb{R}^{N+m}$. If the system is time-varying, historical snapshots are typically discounted in the cost function to prioritize recent data. Hence, the online parameter estimation objective can be framed as minimizing an exponentially weighted least-squares cost
\begin{equation}
\label{eq:ls_objective}
\min_{\bm{\Theta}_k} \sum_{i=1}^{k} \lambda^{k-i} \left\| \bm{\Psi}(\mathbf{x}_{i+1}) - \bm{\Theta}_k \bm{\zeta}_i \right\|_2^2,
\end{equation}
where $\lambda \in (0, 1)$ denotes the exponential forgetting factor (Fig.~\ref{fig:covariance_windup}). At iteration $k$, the estimate $\bm{\Theta}_{k-1}$ is available before
the $k$-th snapshot pair $(\bm{\zeta}_k,\bm{\Psi}(\mathbf{x}_{k+1}))$
is assimilated; hence, the parameter update is driven by the a priori
innovation. Minimizing~\eqref{eq:ls_objective} recursively at each time step yields the generalized \ac{RLS} estimator with covariance adaptation as
\begin{subequations}
\begin{align}
\bm{\epsilon}_k &= \bm{\Psi}(\mathbf{x}_{k+1}) - \bm{\Theta}_{k-1} \bm{\zeta}_k, \label{eq:innovation} \\
\mathbf{K}_k &= \frac{\mathbf{P}_{k-1} \bm{\zeta}_k}{\lambda + \bm{\zeta}_k^\top \mathbf{P}_{k-1} \bm{\zeta}_k}, \label{eq:rls_gain} \\
\bm{\Theta}_k &= \bm{\Theta}_{k-1} + \bm{\epsilon}_k \mathbf{K}_k^\top, \label{eq:param_update} \\
\mathbf{P}_k &= \frac{1}{\lambda} \left( \mathbf{P}_{k-1} - \mathbf{K}_k \bm{\zeta}_k^\top \mathbf{P}_{k-1} \right), \label{eq:cov_update}
\end{align}
\end{subequations}
where $\bm{\epsilon}_k \in \mathbb{R}^N$ represents the a priori innovation sequence within the lifted observable space, $\mathbf{K}_k \in \mathbb{R}^{N+m}$ is the estimator gain vector, and $\mathbf{P}_k \in \mathbb{R}^{(N+m) \times (N+m)}$ denotes the estimation error covariance matrix. The covariance matrix can be initialized using a pre-collected dataset\footnote{It shall contain at least $n$ snapshots to ensure the snapshot matrix has full rank for computing $\bm{\Theta}_k$ and $\mathbf{P}_k$ via, e.g., standard EDMD algorithm.} or as a symmetric positive-definite matrix $\mathbf{P}_0 = \alpha \mathbf{I}$, where $\alpha \in \mathbb{R} \gg 0$ and $\mathbf{I}$ represents the identity matrix of appropriate dimension. By using the recursive updates in~\eqref{eq:innovation}--\eqref{eq:cov_update}, online estimation eliminates the need to store historical snapshot data matrices, requiring only real-time maintenance of $\bm{\Theta}_k$ and $\mathbf{P}_k$ (Fig.~\ref{fig:CR-RKL}).

\subsection{Analysis of Algorithmic and Structural Failure Modes}
\label{subsec:failure_mode}
\revised{Recursive Koopman identification inherits classical \ac{RLS} windup and magnifies it through high-dimensional, collinear dictionaries $\bm{\Psi}(\mathbf{x})$.}
% Recursive system identification with high-dimensional Koopman operators introduces numerical instabilities absent in lower-dimensional adaptive control. These arise from a trade-off between information loss (exponential forgetting) and information gain (directional excitation), with severity governed by the geometry of the lifting dictionary $\bm{\Psi}(\mathbf{x})$. 
Applying the Woodbury identity to~\eqref{eq:cov_update} and unfolding this recurrence from $k=0$ gives
\begin{equation}
\label{eq:info_matrix_step}
\mathbf{P}_k^{-1} = \lambda \mathbf{P}_{k-1}^{-1} + \bm{\zeta}_k \bm{\zeta}_k^\top= \lambda^k \mathbf{P}_0^{-1} + \sum_{i=1}^{k} \lambda^{k-i} \bm{\zeta}_i \bm{\zeta}_i^\top.
\end{equation}

For a recursive estimator without forgetting ($\lambda = 1$), \eqref{eq:info_matrix_step} reduces to a monotonic sum. In this case, even when the robot enters a region with no excitation ($\bm{\zeta}_i = \mathbf{0}$), $\mathbf{P}_k^{-1}$ remains constant, which implies that the error covariance $\mathbf{P}_k$ is bounded from above and does not suffer from numerical windup. However, as informative measurements accumulate, the excited direction of $\mathbf{P}_k^{-1}$ grows monotonically, causing its inverse to tend asymptotically to zero (Fig.~\ref{fig:covariance_windup}(b)). This leads to the well-known estimator freezing phenomenon, in which the adaptation gain vanishes ($\mathbf{K}_k \to \mathbf{0}$). Consequently, the estimator becomes insensitive to time-varying dynamics. 

To maintain responsiveness in the presence of non-stationary dynamics, an exponential forgetting factor $\lambda \in (0, 1)$ shall be included. However, historical information decays exposes weakly excited directions to gain inflation. 
\subsubsection{Unbounded Lifting and Parameter Divergence}
\revised{Globally supported polynomial functions can become badly scaled over long trajectories,} causing the physical state norm and the high‑dimensional regressor vector to grow without bound ($\|\mathbf{x}_k\|_2 \to \infty$, $\|\bm{\zeta}_k\|_2 \to \infty$). Minimizing the objective function~\eqref{eq:ls_objective} under such large coordinate scaling could force the recursive regression to continuously and artificially rescale the parameter rows in $\bm{\Theta}_k$. This artifact of the optimization may lead to \revised{large or physically implausible operator entries}, potentially making the identified linear operators structurally unstable for downstream real‑time \ac{MPC}. \revised{This should be viewed as a risk rather than a universal divergence theorem, and Section~\ref{sec:results} reports one observed case of control failure.}

\subsubsection{Exponential Covariance Windup}
\revised{Localized Gaussian \acp{RBF}, often adopted in \ac{EDMD}, remain bounded but can leave directions weakly excited when the state is far from their centers.} Consider a scenario where a robot enters a low‑entropy operating regime (e.g., steady‑state hovering, cruising between regimes, or following a trajectory confined to a low‑dimensional subspace of $\mathbb{R}^{N+m}$), the RBF evaluations tend to zero in its weakly excited direction over an interval, then information in that direction decays approximately as $\lambda$. Under these under‑excited conditions, the summation term in \eqref{eq:info_matrix_step} no longer counteracts the exponential decay of the regularization history, so the information update reduces to $\mathbf{P}_k^{-1} = \lambda^k \mathbf{P}_0^{-1}$. Taking the limit $k \to \infty$ reveals unbounded growth of the error covariance matrix (Fig.~\ref{fig:covariance_windup}(a)). \revised{Such large $\|\mathbf{P}_k\|$ can amplify innovations into parameter fluctuations, which may destabilize closed‑loop controller and causing system failure.}
\revised{\begin{remark}[Interpretation of Covariance]
    Under the upcoming regulation mechanisms designed to prevent these failure modes, the matrix $\mathbf{P}_k$ weakens in its interpretation as a statistically calibrated covariance of estimation error. Instead, it functionally serves as an adaptation matrix that dictates the step size and direction of parameter updates.
\end{remark}}

\subsection{Robust Covariance-Regulated Modifications}
\label{subsec:strategies}
We introduce three covariance-regulated strategies within the recursive Koopman learning framework (Fig.~\ref{fig:CR-RKL}).

\subsubsection{Error Dead-Zone Gating}
\revised{We propose a novel error dead-zone gating strategy that couples the covariance update directly to the instantaneous predictive ability of the lifted
model~\eqref{eq:lifted_affine}.} Let $\delta > 0$ be a scalar error threshold. We adopt the following conditional update law for the covariance matrix
\begin{equation}
\mathbf{P}_k = \begin{cases} 
\frac{1}{\lambda} \left( \mathbf{P}_{k-1} - \mathbf{K}_k \bm{\zeta}_k^\top \mathbf{P}_{k-1} \right), & \text{if } \|\bm{\epsilon}_k\|_2 > \delta \\
\mathbf{P}_{k-1}, & \text{if } \|\bm{\epsilon}_k\|_2 \le \delta
\end{cases}
\end{equation}
When the norm of the lifted innovation sequence satisfies $\|\bm{\epsilon}_k\|_2 \le \delta$, the identifier assumes that the current operator $\bm{\Theta}_{k-1}$ has sufficient local predictive capability. Under this condition, the covariance update is suspended ($\mathbf{P}_k = \mathbf{P}_{k-1}$), thereby avoiding the exponential inflation factor $\lambda^{-1}$ on gated samples. This gating scheme differs from existing error-threshold methods that completely skip the entire Koopman identification~\cite{calderon2021koopman,loya2026koopman}. Skipping the full update may introduce structural discontinuities and parameter chatter near the dead-zone boundaries while also rejecting small but structurally informative innovations that could track slowly drifting dynamics. In contrast, by freezing the covariance matrix $\mathbf{P}_k$ while continuing to update the parameter matrix~\eqref{eq:param_update}, the identifier operates as a gradient descent optimizer with a stable, non-zero learning rate determined by the historical uncertainty profile stored in $\mathbf{P}_{k-1}$ during uninformative intervals. Hence, this configuration may help isolate high-frequency sensor noise while permitting continuous refinement of the operator estimates.

\subsubsection{Constant-Trace Normalization}
\revised{We adapt the constant-trace normalization from classical adaptive control~\cite{lozano1985globally,landau2011adaptive,airimitoaie2025accelerating}, which is used to relax \ac{PE} requirements. It rescales the intermediate adaptation matrix to a prescribed target $\tau_0$:} 
\begin{equation}
\text{tr}(\mathbf{P}_k) \equiv \text{tr}(\mathbf{P}_0) = \tau_0, \quad \forall k \ge 0.
\end{equation}
At each step, the estimator computes an intermediate adaptation matrix $\mathbf{P}_{\text{temp}}$ via \eqref{eq:cov_update} and its trace $\tau_k = \text{tr}(\mathbf{P}_{\text{temp}})$. The regulated matrix is then obtained by 
\begin{equation}
\label{eq:ct_normalization}
\mathbf{P}_k = \frac{\tau_0}{\tau_k} \mathbf{P}_{\text{temp}}.
\end{equation}
We examine it in high-dimensional Koopman dictionaries. If $\mathbf{P}_\mathrm{temp}\succ0$, scalar rescaling preserves positive definiteness and gives $\lambda_{\max}(\mathbf{P}_k) \leq ||\mathbf{P}_k||_\text{F} \leq \text{tr}(\mathbf{P}_k) = \tau_0$. Thus, no eigenvalue can diverge, and the full matrix cannot converge to zero while $\tau_0>0$. \revised{Moreover, since the previous normalized matrix satisfies $||\mathbf{P}_{k-1}||_2\leq\tau_0$, the next gain follows $||\mathbf{K}_k||_2\leq\tau_0||\bm\zeta_k\||_2/\lambda$, which corresponds to a gain-shaping bound conditional on the regressor magnitude. Notably, while trace evaluation is inexpensive (requiring only $\mathcal{O}(N)$ diagonal additions), it can neither bound the minimum eigenvalue $\lambda_\mathrm{min}(\mathbf{P}_k)$ away from zero nor bound the condition number $\kappa(\mathbf{P}_k)$, unlike SIFt-RLS~\cite{lai2024sift}, which gives a positive lower eigenvalue bound as shown later.}

At each individual step, uniform scalar scaling~\eqref{eq:ct_normalization} exactly preserves the geometric structure of $\mathbf{P}_k$. This is because the normalization operation acts as a homogeneous scalar multiplier $\gamma_k = \tau_0/\tau_k$, therefore it commutes with the spectral decomposition of the intermediate matrix as
$\mathbf{P}_k = \gamma_k \mathbf{P}_{\text{temp}} = \gamma_k \left( \mathbf{V} \mathbf{\Lambda} \mathbf{V}^\top \right) = \mathbf{V} \left( \gamma_k \mathbf{\Lambda} \right) \mathbf{V}^\top$,
where $\mathbf{V}$ is the matrix of eigenvectors and $\mathbf{\Lambda}$ the diagonal matrix of eigenvalues. Hence, the orientation of the principal axes of the covariance ellipsoid remains unchanged. \revised{Therefore, it is better interpreted as an adaptation matrix.}

\subsubsection{\revised{Subspace-of-Information Forgetting}}
\label{app:SIFt}
\revised{Originally, SIFt-RLS forgets information only in the row space of the current
regressor, rather than uniformly discounting all parameter
directions~\cite{lai2024sift}. Let
$\mathbf{R}_{k-1}=\mathbf{P}_{k-1}^{-1}\in\mathbb{R}^{(N+m)\times (N+m)}$,
and let $\varepsilon_{\rm S}>0$ denote the information-filter
threshold. Each row of $\bm{\Theta}_k$ constitutes a scalar-output
regression with the common regressor
$\bm{\zeta}_k^\top\in\mathbb{R}^{1\times (N+m)}$. Consequently, the
general singular-value filter in SIFt-RLS reduces here to retaining
the single information direction when
$\|\bm{\zeta}_k\|_2^2\ge\varepsilon_{\rm S}$. If this condition is
not satisfied, no information direction is retained and set
$\bm{\Theta}_k=\bm{\Theta}_{k-1}$,
$\mathbf{R}_k=\mathbf{R}_{k-1}$, and
$\mathbf{P}_k=\mathbf{P}_{k-1}$. When the direction is retained, define
$\rho_k=\bm{\zeta}_k^\top\mathbf{R}_{k-1}\bm{\zeta}_k>0$.
The resulting rank-one specialization is
\begin{subequations}
\begin{align}
\bar{\mathbf{P}}_{k-1}
&=
\mathbf{P}_{k-1}
+\frac{1-\lambda}{\lambda\rho_k}
\bm{\zeta}_k\bm{\zeta}_k^\top,\\
\mathbf{P}_k
&=
\bar{\mathbf{P}}_{k-1}
-
\frac{
\bar{\mathbf{P}}_{k-1}
\bm{\zeta}_k\bm{\zeta}_k^\top
\bar{\mathbf{P}}_{k-1}}
{1+\bm{\zeta}_k^\top
\bar{\mathbf{P}}_{k-1}\bm{\zeta}_k},\\
\mathbf{R}_k
&=
\mathbf{R}_{k-1}
-\frac{1-\lambda}{\rho_k}
\mathbf{R}_{k-1}
\bm{\zeta}_k\bm{\zeta}_k^\top
\mathbf{R}_{k-1}
+\bm{\zeta}_k\bm{\zeta}_k^\top,\\
\mathbf{K}_k^{\rm S}
&=
\mathbf{P}_k\bm{\zeta}_k,\qquad
\bm{\Theta}_k
=
\bm{\Theta}_{k-1}
+\bm{\epsilon}_k(\mathbf{K}_k^{\rm S})^\top ,
\end{align}
\end{subequations}
where $\bm{\epsilon}_k$ is the a priori innovation defined
in~\eqref{eq:innovation}. 
% The implementation evaluates the dual
% $\mathbf{R}_k$ and $\mathbf{P}_k$ rank-one recursions and
% symmetrizes both matrices to limit round-off drift. 
Because all
lifted outputs share the same regressor, the common gain
$\mathbf{K}_k^{\rm S}$ updates all rows of $\bm{\Theta}_k$
simultaneously.
\begin{remark}[Bounds of Adaptation Matrix~\cite{lai2024sift}]
    For $\mathbf{R}_0\succ\mathbf{0}$, the SIFt-RLS covariance bound
satisfies $\lambda_{\max}(\mathbf{P}_k)
\le
\max\left\{
\frac{1-\lambda}{\varepsilon_{\rm S}},
\lambda_{\max}(\mathbf{P}_0)
\right\}$
without requiring \ac{PE}. If the regressors are
uniformly bounded such that
$\|\bm{\zeta}_k\|_2^2\le\beta$, then $\lambda_{\min}(\mathbf{P}_k)
\ge
\min\left\{
\frac{1-\lambda}{\beta},
\lambda_{\min}(\mathbf{P}_0)
\right\}$.
% These adaptation-matrix bounds do not imply prediction accuracy or
% closed-loop stability for the present uncertain, time-varying
% Koopman problem. 
% In particular, the parameter-error stability
% results in assume a fixed, noise-free regression
% model and additionally require \ac{PE} for global
% uniform exponential stability. 
\end{remark}}

\section{Case Studies}
\label{sec:case_study}

\begin{figure}
    \centering
    \includegraphics[width=\linewidth]{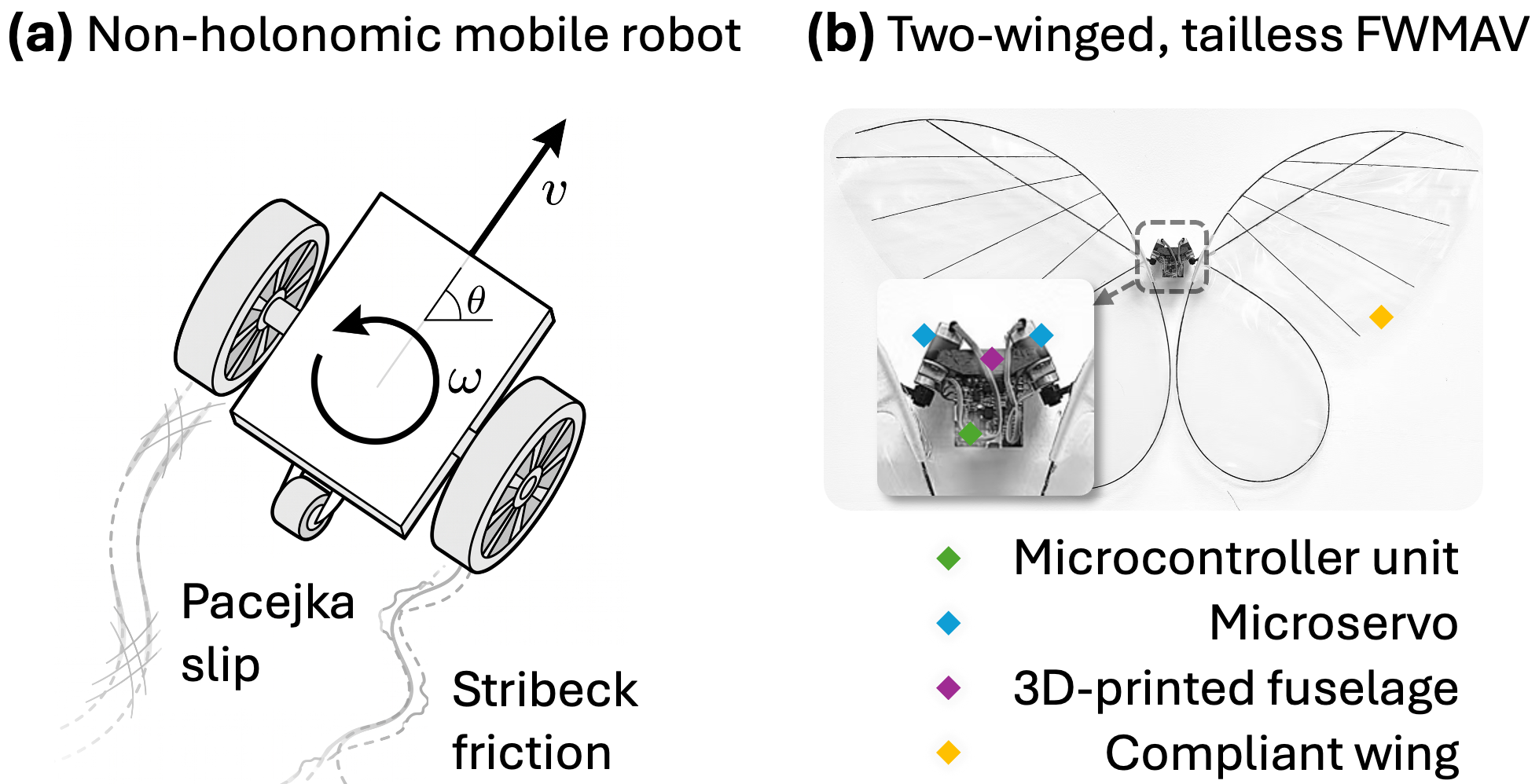}
    \caption{\textbf{Two case studies for evaluating CR-RKL.} \textbf{(a)} A simulated DDR under Pacejka slip and Stribeck friction. \textbf{(b)} A butterfly-inspired FWMAV, exhibiting highly nonlinear aerodynamics.}
    \label{fig:robot}
    \vspace{-1em}
\end{figure}

We consider a synthetic non-holonomic robot with scheduled
vehicle–terrain interactions (Fig.~\ref{fig:robot}(a)) and an experimental flight dataset from an untethered 26-g two-winged, tailless \ac{FWMAV} (Fig.~\ref{fig:robot}(b)). In both cases the online estimation receives states and inputs but not the scheduled slip/friction parameters or flight-regime label.

\subsection{Case Study I: Simulated Non-Holonomic Mobile Robot}
\label{subsec:case_study_1}
The first validation platform is a \ac{DDR} operating under
non-stationary physical conditions where tractive degradation and state-dependent nonlinear disturbances corrupt the
nominal kinematic assumptions.

\subsubsection{Nominal Kinematics}
The state vector of the vehicle relative to an inertial reference frame is defined as $\mathbf{x}_k = [p_{x,k}, p_{y,k}, \theta_k]^\top \in \mathbb{R}^3$, where $p_{x,k}$ and $p_{y,k}$ denote the planar Cartesian positions and $\theta_k$ denotes the heading orientation. Driven by a control input vector $\mathbf{u}_k = [v_k, \omega_k]^\top \in \mathbb{R}^2$ containing the nominal forward linear velocity and rotational yaw rate, respectively, the discrete-time nominal kinematics propagate at a loop interval of $\Delta t = 0.05\,\text{s}$ according to
\begin{subequations}    
\begin{align}
p_{x,k+1} &= p_{x,k} + v_k \cos(\theta_k)\Delta t, \\
p_{y,k+1} &= p_{y,k} + v_k \sin(\theta_k)\Delta t, \\
\theta_{k+1} &= \theta_k + \omega_k \Delta t.
\end{align}
\end{subequations}
% To evaluate long-horizon parametric alertness and closed-loop control under persistent unmodeled variations, 
To evaluate modeling over a long horizon, the robot is driven along a trajectory via periodic open-loop commands: $v_k = 1.5 + 0.3\sin(0.1k)$ and $\omega_k = 0.6\cos(0.08k)$.

\subsubsection{Non-Stationary Disturbance and Tractive Slip Dynamics}
To emulate challenging field deployment conditions, the physical plant simulation is augmented with a highly nonlinear, time-varying interaction model that represents wheel slip, lateral skidding, and velocity-dependent body drag: (i) Pacejka-based motion slip: Tractive velocity degradation is introduced by evaluating an instantaneous slip proxy $s_k = 0.18|v_k\omega_k|$ through a localized Pacejka Magic Formula curve~\cite{pacejka2006tyre}. This yields a nonlinear motion attenuation factor $\ell_k \in [0, 1]$ that down-scales the effective forward speed $\tilde{v}_k$ and yaw rate $\tilde{\omega}_k$, while generating an orthogonal lateral skidding velocity component $v_{\perp,k} = \operatorname{sign}(\omega_k) \cdot 0.45 v_k \ell_k$. (ii) Velocity-dependent Stribeck friction: The vehicle encounters non-conservative body-centric resistive forces modeled via an empirical Stribeck composition $\mu(v_{\text{body},k})$ along both the longitudinal and transverse axes. This friction introduces static, kinetic Coulomb, and viscous drag transitions that distort the linear mapping between inputs and states. (iii) Multi-phase scheduling: To assess the modeling and control performance of the identified operator under unmodeled physical variations, these hidden dynamics are modulated over a continuous timeline across five sequential phases: nominal kinematics, Pacejka wheel slip, Stribeck friction transitions, combined multi-source uncertainties, and a recovery back to nominal kinematics.

\subsection{Case Study II: Flapping-Wing Micro Aerial Vehicle}
\label{subsec:case_study_2}

The second benchmark uses real-flight data collected from
an \ac{FWMAV} operating untethered indoors.

\subsubsection{System State Space and Control Input Configuration}
The 100-Hz telemetry provides ten channels: three gyroscope 
channels $\mathbf{g}_k \in \mathbb{R}^3$, three accelerometer channels 
$\mathbf{a}_k \in \mathbb{R}^3$, pitch, roll, and yaw estimates 
$(\theta_k, \phi_k, \psi_k)$, and a yaw-rate estimate $\dot{\psi}_k$. 
Thus $\mathbf{x}_k = [g_k^\top, a_k^\top, \theta_k, \phi_k, \psi_k, \dot{\psi}_k]^\top \in \mathbb{R}^{10}$. 
Each maneuver dataset has one scalar control channel: $u_k$ is the 
pitch-offset command for pitch maneuvers and the yaw-offset command 
for yaw maneuvers. These commands modulate the wing actuation of the 
platform described in~\cite{gu202626}.

\subsubsection{Cross-Regime Out-of-Domain Generalization}
To test online adaptation across distinct operating regimes, flight trajectories are split into pitch and yaw maneuvers. The Koopman operator is initialized offline using only the source regime (e.g., climbing) and then updated online during the target regime (e.g., turning). This cross-regime setup explicitly tests whether covariance-regulated updates can adapt to unseen, non-stationary aerodynamics without triggering covariance windup.

\section{Results and Discussion}
\label{sec:results}

\subsection{Comparison Protocol}
\label{subsec:comparison_protocol}

\begin{table}[t]
\centering
\footnotesize
\setlength{\tabcolsep}{4.5pt}
\renewcommand{\arraystretch}{0.92}
\revised{\caption{Classification of Online Update Rules for Comparison}
\label{tab:method_taxonomy}
}
\begin{tabular}{llll}
\toprule
\textbf{Name/Suffix} & \textbf{Dictionary} & \textbf{Updating $\bm{\Theta}$} & \textbf{Updating $\mathbf{P}$} \\
\midrule
Offline RBF & $\texttt{RBF}$ & \cellcolor{gray!20}{None online} & \cellcolor{gray!20}{None online} \\
Vanilla & $\texttt{Poly}$ / $\texttt{RBF}$ & \cellcolor{orange!20}{Every sample} & \cellcolor{orange!20}{Exponential forgetting} \\
DZ-CS & $\texttt{Poly}$ / $\texttt{RBF}$ & \cellcolor{orange!20}{Every sample} & \cellcolor{cyan!20}{Freeze if gated} \\
DZ-KS & $\texttt{Poly}$ / $\texttt{RBF}$ & \cellcolor{cyan!20}{Freeze if gated} & \cellcolor{cyan!20}{Freeze if gated} \\
CT & $\texttt{Poly}$ / $\texttt{RBF}$ & \cellcolor{orange!20}{Every sample} & \cellcolor{orange!20}{Update (set trace to $\tau_0$)} \\
SIFt & $\texttt{Poly}$ / $\texttt{RBF}$ & \cellcolor{orange!20}{Every sample} & \cellcolor{orange!20}{Directional forgetting} \\
\bottomrule
\end{tabular}
\vspace{-1em}
\end{table}

\textbf{Dictionary selection.} Two common types of dictionaries were studied. The polynomial dictionary ($\texttt{Poly}$) is defined as $\bm{\Psi}_\text{poly}(\mathbf{x})=[p_x, p_y, \theta, p_x^2, p_y^2, \sin\theta, \cos\theta]^\top$, which provides global unbounded support. The Gaussian \ac{RBF} dictionary ($\texttt{RBF}$) is defined as $\bm{\Psi}_\text{RBF}(\mathbf{x}) = [p_x, p_y, \theta, \phi_1(\mathbf{p}), \dots, \phi_{16}(\mathbf{p}), \sin\theta, \cos\theta, \text{atan2}(\sin\theta,\cos\theta)]^\top$, where $\phi_i(\mathbf{p}) = \exp(-\|\mathbf{p}-\mathbf{c}_i\|_2^2/2\sigma^2)$ are localized kernels with grid centers $\mathbf{c}_i \in \mathbb{R}^2$. 

\textbf{Comparative classification.} Eleven methods were evaluated \revised{(Table~\ref{tab:method_taxonomy})}: an offline \ac{RBF} baseline, vanilla online variants, proposed robust variants incorporating error dead-zone gating (``DZ'') and constant-trace normalization (``CT''), and \revised{a variant with directional forgetting (``SIFt'')}. The dead-zone strategy is further categorized as our proposed covariance skipping (``CS'') or conventional full Koopman update skipping (``KS''). \revised{These methods were used consistently across every \ac{DDR} identification and control experiment, as well as in the four flight-regime-transfer directions for \ac{FWMAV}.}

\textbf{Hyperparameter setting and performance metrics.} \revised{All online errors were computed from the pre-update prediction $\bm{\Theta}_{k-1}\bm{\zeta}_k$. Unless explicitly stated, DDR uses $\lambda=0.98$, $\mathbf{P}_0=100\mathbf{I}$, $\delta=10^{-2}$, $\tau_0=\operatorname{tr}(\mathbf{P}_0)$, and $\varepsilon_{\rm S}=10^{-10}$; flight experiments use $\lambda=0.995$, $\mathbf{P}_0=100\mathbf{I}$, $\delta=1$, and $\varepsilon_{\rm S}=10^{-10}$. Dead-zone thresholds are reported as experiment-specific settings rather than universal constants.} Performance is quantified via the mean single-step prediction error $\bar{e}_\mathbf{x} = \frac{1}{N} \sum_{k=0}^{N-1} \|\mathbf{x}_{k+1} - \hat{\mathbf{x}}_{k+1}\|_2$, along with the Frobenius norms of the covariance matrix and the Koopman transition matrix. \revised{Throughout the reported tables, we use color gradients to aid interpretation: orange shading highlights larger numerical values (with darker shades indicating worse modeling or tracking performance), whereas teal shading highlights smaller numerical values (with darker shades indicating more numerically stable performance).}

\subsection{One-Step Learning Performance for DDR}
\label{subsec:results_ddr}

\begin{table}[!t]
\centering
\scriptsize
\setlength{\tabcolsep}{10pt}
\renewcommand{\arraystretch}{0.8}
\begin{threeparttable}
\caption{Learning Performance of a DDR under Unmodeled Dynamics}
\label{tab:ddr_online_modeling_results}
\begin{tabular}{lccc}
\toprule
\textbf{Method} & \textbf{Mean Error} $\mathbf{\bar{e}_{\mathbf{x}}}$ & \textbf{Final }$\|\mathbf{P}\|_\text{F}$ & \textbf{Final }$\|\mathbf{A}\|_\text{F}$ \\ \midrule
Offline RBF & \cellcolor{reddark}{$1.30 \times 10^{-2}$} & -- & \cellcolor{greendark}{$7.62 \times 10^1$} \\
Poly & \cellcolor{redlight}{$1.96 \times 10^{-3}$} & \cellcolor{greenmed}{$1.94 \times 10^4$} & \cellcolor{greendark}{$5.32 \times 10^1$} \\
Poly (DZ-CS) & \cellcolor{redlight}{$1.95 \times 10^{-3}$} & \cellcolor{greenmed}{$1.95 \times 10^4$\tnote{$\dagger$}} & \cellcolor{greendark}{$5.31 \times 10^1$} \\
Poly (DZ-KS) & \cellcolor{redlight}{$1.98 \times 10^{-3}$} & \cellcolor{greenmed}{$1.94 \times 10^4$\tnote{$\dagger$}} & \cellcolor{greendark}{$5.29 \times 10^1$} \\
Poly (CT) & \cellcolor{redmed}{$4.47 \times 10^{-3}$} & \cellcolor{greendark}{$9.00 \times 10^2$} & \cellcolor{greenmed}{$6.37 \times 10^2$} \\
\revised{Poly (SIFt)} & \cellcolor{redmed}{$4.24 \times 10^{-3}$} & \cellcolor{greenverydark}{$6.45 \times 10^1$} & \cellcolor{greendark}{$8.87 \times 10^1$} \\
\revised{RBF} & \cellcolor{redverydark}{$1.95 \times 10^{0}$} & \cellcolor{greenlight}{$2.51 \times 10^{46}$} & \cellcolor{greenlight}{$3.93 \times 10^4$} \\
RBF (DZ-CS) & \cellcolor{redlight}{$1.08 \times 10^{-3}$} & \cellcolor{greenmed}{$3.47 \times 10^3$\tnote{$\ddagger$}} & \cellcolor{greenverydark}{$2.93 \times 10^0$} \\
RBF (DZ-KS) & \cellcolor{redmed}{$4.64 \times 10^{-3}$} & \cellcolor{greenmed}{$2.39 \times 10^4$\tnote{$\ddagger$}} & \cellcolor{greenverydark}{$2.97 \times 10^0$} \\
RBF (CT) & \cellcolor{redmed}{$5.57 \times 10^{-3}$} & \cellcolor{greendark}{$6.64 \times 10^2$} & \cellcolor{greenverydark}{$2.98 \times 10^0$} \\
\revised{RBF (SIFt)} & \cellcolor{redmed}{$4.19 \times 10^{-3}$} & \cellcolor{greendark}{$3.48 \times 10^2$} & \cellcolor{greenverydark}{$2.92 \times 10^0$} \\
\bottomrule
\end{tabular}
\begin{tablenotes}
\footnotesize
% \item[$\dagger$] DZ-CS: dead-zone with covariance skipping; DZ-KS: dead-zone with full Koopman skipping; CT: constant-trace normalization.
\item[$\dagger$] Covariance/full-update skip rate: 3.24\% (DZ-CS) / 3.02\% (DZ-KS)
\item[$\ddagger$] Covariance/full-update skip rate: 97.74\% (DZ-CS) / 95.8\% (DZ-KS)
\end{tablenotes}
\end{threeparttable}
\vspace{-2em}
\end{table}

\begin{figure}
    \centering
    \includegraphics[width=\linewidth]{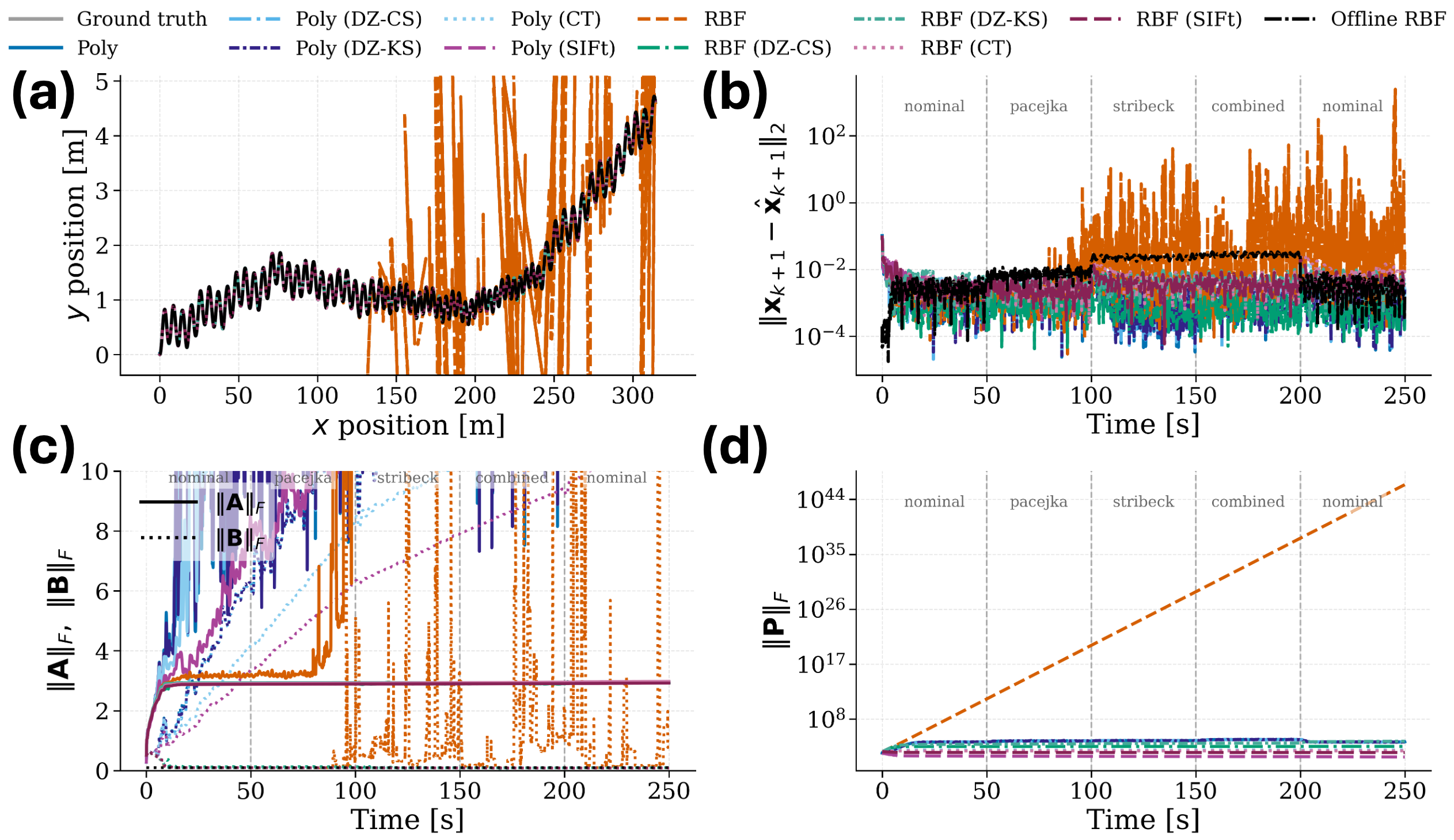}
    \caption{\textbf{Online Koopman learning for a DDR under unmodeled dynamic transitions.} \textbf{(a)} Predicted and true planar paths. \textbf{(b)} Pre-update one-step physical-state error. \textbf{(c)} Learned operator norms $||\mathbf{A}_k||_\mathrm{F}$ (solid) and $||\mathbf{B}_k||_\mathrm{F}$ (dotted). \textbf{(d)} Adaptation matrix norm $||\mathbf{P}_k||_\mathrm{F}$. Vertical lines identify the five scheduled dynamics phases.}
    \label{fig:DDR_online_modeling_benchmark}
\end{figure}

We first evaluate online adaptation and gain regulation on the 250-second \ac{DDR} trajectory comprising $N=5000$ time steps (Section~\ref{subsec:case_study_1}). Table~\ref{tab:ddr_online_modeling_results} and Fig.~\ref{fig:DDR_online_modeling_benchmark} report the mean error, final adaptation norms, and operator norms.

\revised{\textbf{Online adaptation improves one-step prediction unless the recursion becomes unstable.}} Excluding vanilla \ac{RBF}, every numerically usable online method improves the offline baseline. Vanilla \ac{RBF} instead exhibits an explosive growth of $\|\mathbf{P}_k\|_\text{F}$ and divergence of $\|\mathbf{A}_k\|_\text{F}$ due to regressor collapse, sometimes even causing $\mathbf{P}_k$ to lose positive-definiteness. Compared to static baseline, recursive update is therefore beneficial under the scheduled dynamical changes when the estimator remains numerically usable.

\revised{\textbf{Covariance skipping is preferable to full Koopman update skipping in every matched comparison.}} With the \texttt{RBF} dictionary, DZ-CS attains an error 76.7\% below DZ-KS at comparable 97.74\%/95.8\% skip rates. The \texttt{Poly} dictionary gain is only 1.5\% with gating occurring on 3.02--3.24\% of samples. This may be attributed to the less structural sparsity from unbounded polynomial features. Since dead-zone rarely engages, it in essence degenerates to the baseline. With threshold held fixed, the paired ablation supports retaining operator correction on gated samples.

\revised{\textbf{Covariance regulation is a failure-mode safeguard rather than a universal accuracy enhancer.} Methods including DZ-CS, DZ-KS, CT, and SIFt prevent the windup issue of vanilla RBF, but can also bias or delay adaptation when vanilla recursive Koopman learning remains usable. For example, CT and SIFt with the \texttt{Poly} dictionary are less accurate than vanilla Poly. Hence, we conclude that the benefit of covariance regulation is numerical survival under weak excitation, not guaranteed prediction performance improvement on every dataset.}

\begin{figure}
    \centering
    \includegraphics[width=\linewidth]{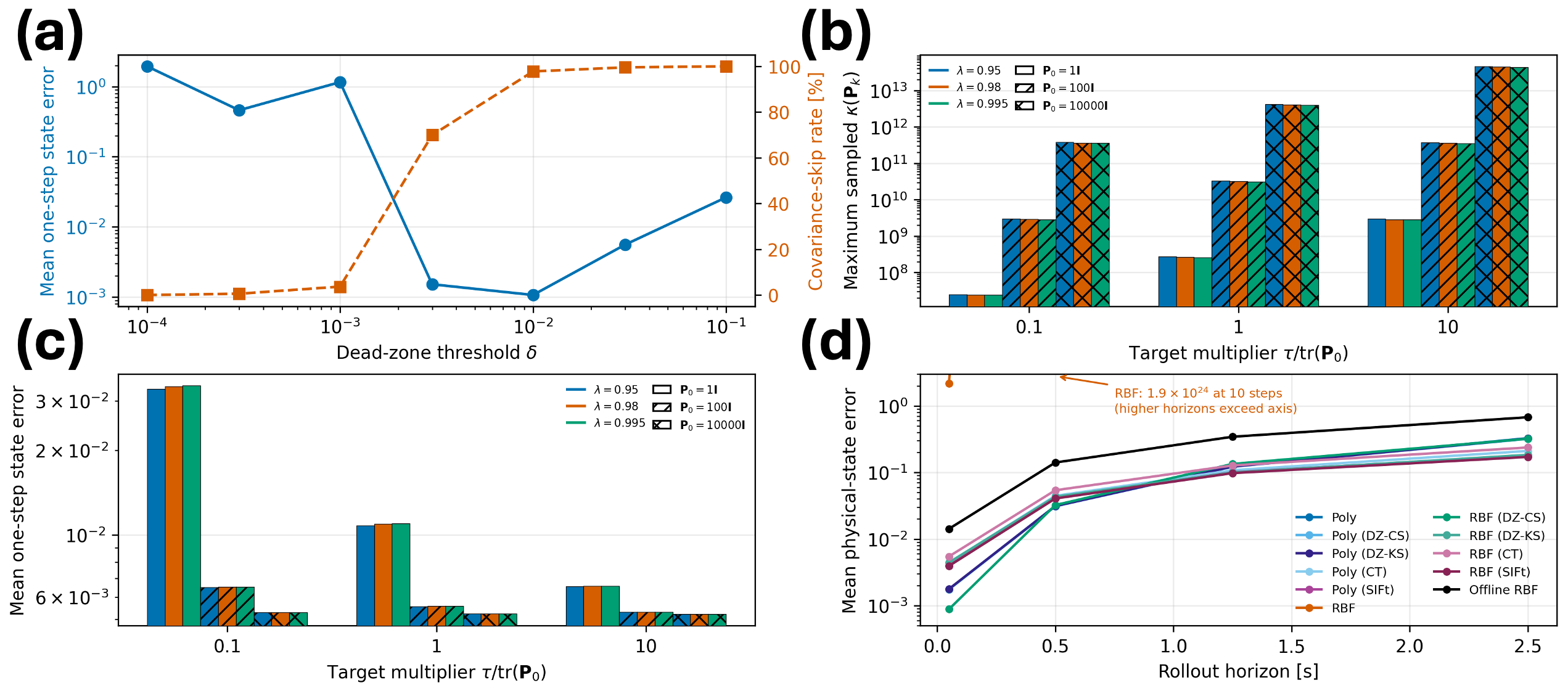}
    \caption{\revised{\textbf{Sensitivity and multi-step diagnostics.} \textbf{(a)} Dead-zone threshold versus one-step error and covariance-skip rate for \texttt{RBF} dictionary. \textbf{(b, c)} Ablation study across 27 settings for constant-trace normalization experiment, shown as grouped bars on a logarithmic scale. Color denotes $\lambda$, hatching denotes $\mathbf{P}_0$, and each horizontal group is one target-trace multiplier. \textbf{(d)} Pre-update frozen-parameter open-loop rollouts for every method in Table~\ref{tab:method_taxonomy}. Each window saves $\bm\Theta_{k-1}$, propagates it without updates using future input sequence, and is scored against future states.}}
    \label{fig:sensitivity}
    \vspace{-1em}
\end{figure}

\revised{\begin{remark}(Hyperparameter Discussion)
    First, we analyzed the sensitivity to the dead-zone threshold $\delta$ (Fig.~\ref{fig:sensitivity}(a)). For \texttt{RBF} dictionary, $\delta=0.003/0.01/0.03$ yield 69.90/97.73/99.50\% skipping and errors 1.52/1.07/5.58$\times 10^{-3}$, whereas $\delta=0.01$ skips only 3.24\% for \texttt{Poly}. Hence, a fixed raw innovation threshold is not directly transferable across dictionaries. Second, we did an ablation study over 27 settings for constant-trace normalization. We considered $\lambda\in\{0.95, 0.98, 0.995\}$ (fast/nominal/slow forgetting), $\mathbf{P}_0\in\{\mathbf{I}, 100\mathbf{I}, 10^4\mathbf{I}\}$, and $\frac{\tau}{\mathrm{tr}(\mathbf{P}_0)}\in\{0.1,1,10\}$ (contraction/preservation/expansion). It is shown that larger target trace often lowers error while worsening conditioning (Fig.~\ref{fig:sensitivity}(b) and (c)).
\end{remark}}

\subsection{Cross-Regime Prediction for Real-Flight FWMAV}

\begin{table*}[t]
\centering
\scriptsize
\setlength{\tabcolsep}{4.2pt}
\renewcommand{\arraystretch}{0.92}

\begin{minipage}[t]{0.62\linewidth}
\centering
\caption{Learning Performance of an FWMAV Across Flight Regimes}
\label{tab:transfer_results}
\begin{tabular}{lcccccc}
\toprule
\textbf{Method} & \multicolumn{4}{c}{\textbf{Mean State Error $\bar e_{\mathbf{x}}$}} & \textbf{Max. Final} $\|\mathbf{P}\|_\text{F}$ & \textbf{Skip-Rate Range} \\
\cmidrule(lr){2-5}
 & \textbf{P$\to$P} & \textbf{Y$\to$Y} & \textbf{P$\to$Y} & \textbf{Y$\to$P} & \textbf{(Four Directions)} & \textbf{(\%)} \\
\midrule
Offline RBF & \cellcolor{redmed}{0.453} & \cellcolor{reddark}{0.595} & \cellcolor{redverydark}{0.889} & \cellcolor{redverydark}{0.862} & -- & -- \\
Poly & \cellcolor{redmed}{0.516} & \cellcolor{reddark}{0.609} & \cellcolor{reddark}{0.629} & \cellcolor{redmed}{0.552} & \cellcolor{greenlight}{$5.81\times10^4$} & -- \\
Poly (DZ-CS) & \cellcolor{reddark}{0.538} & \cellcolor{reddark}{0.627} & \cellcolor{reddark}{0.644} & \cellcolor{reddark}{0.591} & \cellcolor{greenmed}{$1.81\times10^3$} & 63.9--73.2 \\
Poly (DZ-KS) & \cellcolor{redverydark}{0.587} & \cellcolor{redverydark}{0.660} & \cellcolor{redverydark}{0.688} & \cellcolor{reddark}{0.621} & \cellcolor{greenmed}{$2.18\times10^3$} & 59.0--69.8 \\
Poly (CT) & \cellcolor{redmed}{0.508} & \cellcolor{reddark}{0.620} & \cellcolor{reddark}{0.641} & \cellcolor{redmed}{0.559} & \cellcolor{greendark}{$1.04\times10^3$} & -- \\
\revised{Poly (SIFt)} & \cellcolor{redmed}{0.508} & \cellcolor{reddark}{0.618} & \cellcolor{reddark}{0.652} & \cellcolor{redmed}{0.568} & \cellcolor{greenverydark}{$3.79\times10^2$} & -- \\
RBF & \cellcolor{redmed}{0.474} & \cellcolor{redmed}{0.565} & \cellcolor{redmed}{0.583} & \cellcolor{redlight}{0.508} & \cellcolor{greenlight}{$9.69\times10^4$} & -- \\
RBF (DZ-CS) & \cellcolor{redmed}{0.500} & \cellcolor{redmed}{0.574} & \cellcolor{reddark}{0.601} & \cellcolor{redmed}{0.530} & \cellcolor{greenmed}{$4.40\times10^3$} & 57.2--70.6 \\
RBF (DZ-KS) & \cellcolor{redmed}{0.527} & \cellcolor{reddark}{0.603} & \cellcolor{reddark}{0.627} & \cellcolor{redmed}{0.565} & \cellcolor{greenmed}{$5.31\times10^3$} & 53.6--67.5 \\
RBF (CT) & \cellcolor{redlight}{0.465} & \cellcolor{redmed}{0.576} & \cellcolor{reddark}{0.597} & \cellcolor{redlight}{0.505} & \cellcolor{greendark}{$1.25\times10^3$} & -- \\
\revised{RBF (SIFt)} & \cellcolor{redlight}{0.464} & \cellcolor{redmed}{0.578} & \cellcolor{reddark}{0.599} & \cellcolor{redlight}{0.508} & \cellcolor{greenverydark}{$6.72\times10^2$} & -- \\
\bottomrule
\end{tabular}
\end{minipage}
\hfill
\begin{minipage}[t]{0.37\linewidth}
\centering
\caption{DDR Control Performance with Learning}
\label{tab:tracking_energy}
\setlength{\tabcolsep}{3pt}
\renewcommand{\arraystretch}{0.76}
\begin{tabular}{lccc}
\toprule
\textbf{Method} & \textbf{Mean Error} & \textbf{Control Energy} & \textbf{Fallback} \\
\midrule
Offline RBF & \cellcolor{redlight}{$4.46 \times 10^{-2}$} & $3.53 \times 10^{0}$ & \cellcolor{greenverydark}{$0.00\%$} \\
Poly & \cellcolor{redmed}{$1.95 \times 10^{-1}$} & $3.91 \times 10^{0}$ & \cellcolor{greenmed}{$0.90\%$} \\
Poly (DZ-CS) & \cellcolor{redverydark}{$3.63 \times 10^{1}$} & $3.43 \times 10^{0}$ & \cellcolor{greenmed}{$0.45\%$} \\
Poly (DZ-KS) & \cellcolor{reddark}{$9.70 \times 10^{0}$} & $6.52 \times 10^{0}$ & \cellcolor{greenmed}{$0.30\%$} \\
Poly (CT) & \cellcolor{redverydark}{$7.02 \times 10^{1}$} & $3.34 \times 10^{-1}$ & \cellcolor{greenlight}{$2.65\%$} \\
\revised{Poly (SIFt)} & \cellcolor{redmed}{$1.36 \times 10^{-1}$} & $3.78 \times 10^{0}$ & \cellcolor{greenmed}{$0.85\%$} \\
RBF & \cellcolor{redverydark}{$7.31 \times 10^{1}$} & $1.19 \times 10^{0}$ & \cellcolor{redverydark}{$100.00\%$} \\
RBF (DZ-CS) & \cellcolor{redlight}{$3.85 \times 10^{-2}$} & $3.57 \times 10^{0}$ & \cellcolor{greenverydark}{$0.00\%$} \\
RBF (DZ-KS) & \cellcolor{redlight}{$3.93 \times 10^{-2}$} & $3.55 \times 10^{0}$ & \cellcolor{greenverydark}{$0.00\%$} \\
RBF (CT) & \cellcolor{redlight}{$4.26 \times 10^{-2}$} & $3.54 \times 10^{0}$ & \cellcolor{greenverydark}{$0.00\%$} \\
\revised{RBF (SIFt)} & \cellcolor{redlight}{$3.90 \times 10^{-2}$} & $3.54 \times 10^{0}$ & \cellcolor{greenverydark}{$0.00\%$} \\
\bottomrule
\end{tabular}
\vspace{1pt}\parbox{0.98\linewidth}{\scriptsize Fallback reports the fraction of steps at which all attempted CVXPY solvers failed and dense solve was hence used. %and the repository used its dense solve plus first-input clipping. A nonzero value means the row is not a pure constrained-QP result.
}
\end{minipage}
\vspace{-1em}
\end{table*}

We next evaluate offline-to-online transfer of a two-winged, tailless \ac{FWMAV} (Section~\ref{subsec:case_study_2}) across pitch and yaw flight regimes. An initial Koopman model is trained offline on a specific source flight regime (e.g., pitch maneuvers) \revised{using 20\% source-regime samples} and evaluated via single-step prediction across both in-domain and out-of-distribution target envelopes (e.g., yaw maneuvers). Quantitative results across four transfer directions are reported in Table~\ref{tab:transfer_results}.% and Fig.~\ref{fig:FWMAV_online_modeling_benchmark}.

\revised{\textbf{Online adaptation is most consistently beneficial under cross-regime shift.}} Every online method improves the offline baseline for pitch-to-yaw (P$\to$Y) and yaw-to-pitch (Y$\to$P) transfer by 22.6--34.4\% and 27.9--41.4\%. In-domain, offline \ac{RBF} is best for pitch-to-pitch (P$\to$P) transfer, while four \ac{RBF} variants except DZ-KS modestly improve yaw-to-yaw (Y$\to$Y) transfer. The benefit is therefore most apparent when the source operator is mismatched.

\revised{\textbf{DZ-CS retains an empirical advantage over DZ-KS, but no regulated method is uniformly best.} DZ-CS is lower in all eight flight comparisons by 0.030--0.049 (\texttt{Poly}) and 0.026--0.035 (\texttt{RBF}) at comparable skip ranges. Cross-regime, however, CT has the lowest regulated-method average for both dictionaries, followed closely by SIFt and DZ-CS. The matched ablation favors covariance-only skipping, while excitation and hyperparameters determine the overall ranking.}

\revised{\begin{remark}(Dead-Zone Threshold Selection)
    $\delta=1$ yields 53.6--73.2\% flight skip rates, whereas $\delta=10^{-2}$ yields about 3\% for \texttt{Poly} and 96\% for \texttt{RBF} on the \ac{DDR}. This confirms platform dependence. To examine a target-independent alternative to manually specifying
$\delta$, we propose a source-calibration approach. Let ``$\mathrm{src}$''
denote the source flight regime (``P'' for pitch or ``Y'' for yaw), let ``$\mathrm{dict}$'' denote the lifting
dictionary (\texttt{Poly} or \texttt{RBF}), and let $\widehat{\bm{\Theta}}_{\mathrm{src},\mathrm{dict}}$ be identified
using the source fitting set. For each sample in a disjoint source
calibration set $\mathcal{I}_{\mathrm{src}}^{\rm cal}$, we compute $r_{j}^{(\mathrm{src},\mathrm{dict})}
=
\left\|
\bm{\Psi}_{\mathrm{dict}}(\mathbf{x}^{\mathrm{src}}_{j+1})
-
\widehat{\bm{\Theta}}_{\mathrm{src},\mathrm{dict}}
\bm{\zeta}^{(\mathrm{src},\mathrm{dict})}_{j}
\right\|_2,\,
j\in\mathcal{I}_{\mathrm{src}}^{\rm cal}$,
where
$\bm{\zeta}^{(\mathrm{src},\mathrm{dict})}_{j}
=
[\bm{\Psi}_{\mathrm{dict}}(\mathbf{x}^{\mathrm{src}}_{j})^\top\;
(\mathbf{u}^{\mathrm{src}}_{j})^\top]^\top$. Let
$r^{(\mathrm{src},\mathrm{dict})}_{1}\leq\cdots\leq r^{(\mathrm{src},\mathrm{dict})}_{n_{\rm cal}}$ denote the ordered calibration scores, where $n_{\rm cal}=|\mathcal{I}_{\rm src}^{\rm cal}|$. The calibrated threshold
$\delta_{\eta}^{(\mathrm{src},\mathrm{dict})}$ is the linearly interpolated empirical
$\eta$-quantile of these source-calibration innovations, which is defined by $h=(n_{\rm cal}-1)\eta,\,
\ell=\lfloor h\rfloor,\,
\gamma=h-\ell,\,
\delta_\eta
=
(1-\gamma)r^{(\mathrm{src},\mathrm{dict})}_{\ell+1}
+
\gamma r^{(\mathrm{src},\mathrm{dict})}_{\ell+2}$. Taking
$\eta=0.75$ as an example, $\delta_{0.75}^{(\mathrm{P},\texttt{Poly})} = 0.850$, $\delta_{0.75}^{(\mathrm{P},\texttt{RBF})} = 1.173$, $\delta_{0.75}^{(\mathrm{Y},\texttt{Poly})} = 1.023$, and $\delta_{0.75}^{(\mathrm{Y},\texttt{RBF})} = 1.203$.
Each calibrated value is held fixed during the corresponding target
adaptation and is reused for both target regimes sharing the same source
regime. Note that $\eta$ specifies a percentile
of the source-calibration innovation distribution but does not prescribe the covariance-update skip rate under a shifted target
distribution.
\end{remark}}

\begin{figure}[ht]
    \centering
    \includegraphics[width=\linewidth]{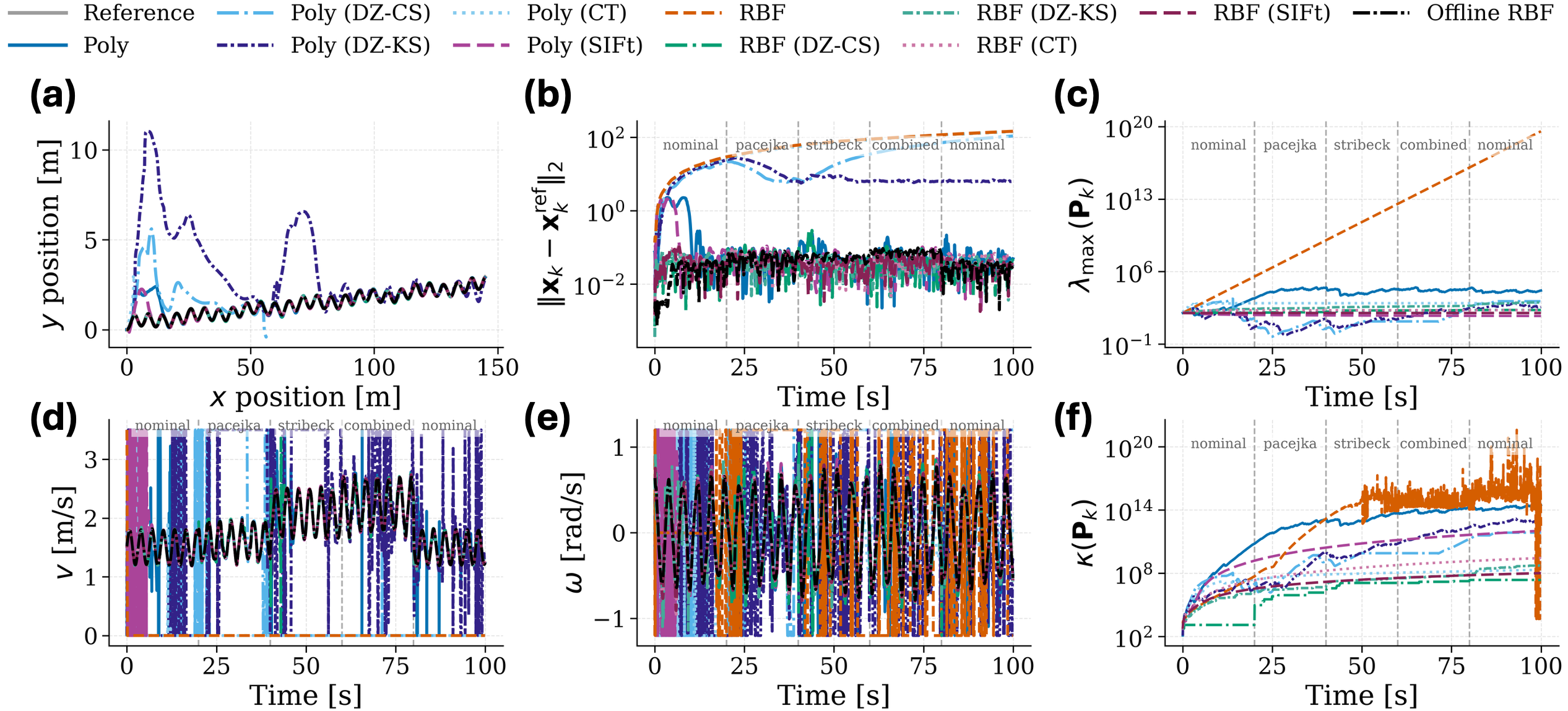}
    \caption{\textbf{Koopman--MPC tracking and adaptation matrix evolution on the DDR.} \textbf{(a)} Predicted and true planar paths. \textbf{(b)} Pre-update one-step physical-state error. \textbf{(d)} Applied forward control input. \textbf{(e)} Applied angular control input. \textbf{(c)} Maximum eigenvalue of adaptation matrix $\lambda_{\max}(\mathbf{P})$. \textbf{(f)} Condition number $\kappa(\mathbf{P}) = \lambda_{\max}(\mathbf{P}) / \lambda_{\min}(\mathbf{P})$, where $\lambda_{\min}(\mathbf{P})=\lambda_{\min}^+(\mathbf{P})$ is the minimum positive eigenvalue for the vanilla RBF method.}
    \label{fig:DDR_control_benchmark}
    \vspace{-1em}
\end{figure}

\subsection{Koopman--MPC with Online Learning for DDR}

\revised{Before closing the loop, we assessed the multi-step prediction performance of each method and found that one-step prediction ranking does not persist with multi-step horizon (Fig.~\ref{fig:sensitivity}(d)). For example, \texttt{RBF} (DZ-CS) grows from $8.84\times 10^{-4}$ at one step to 0.321 at 50 steps, whereas \texttt{RBF} (SIFt) grows from $3.94\times 10^{-3}$ to 0.171. Likewise, \texttt{Poly} (SIFt) gives the best 50-step error (0.180 vs. 0.320).}

We then evaluate closed-loop tracking under non-stationary dynamics with the online Koopman models embedded in an \ac{MPC}. The Koopman--MPC formulation uses a prediction and control horizon of $N_p = N_u = 50$ steps ($2.5\,\text{s}$). At each sampling instant, the following \ac{QP} is solved:
\begin{subequations}
\label{eq:mpc_problem}
\begin{align}
\min_{\mathbf{u}_0, \dots, \mathbf{u}_{N_p-1}} \quad & \sum_{i=1}^{N_p} \| \hat{\mathbf{x}}_{k+i} - \mathbf{x}^{\mathrm{ref}}_{k+i} \|_{\mathbf{Q}}^2 + \sum_{i=0}^{N_p-1} \| \mathbf{u}_{k+i} \|_{\mathbf{R}}^2 \\
\text{s.t.} \quad & \mathbf{z}_0 = \bm{\Psi}(\mathbf{x}_k), \\
& \mathbf{z}_{i+1} = \mathbf{A}_k \mathbf{z}_i + \mathbf{B}_k \mathbf{u}_i, \\
& \hat{\mathbf{x}}_{k+i} = \mathbf{C} \mathbf{z}_{i}, \\
& \mathbf{u}_{\min} \le \mathbf{u}_{k+i} \le \mathbf{u}_{\max},
\end{align}
\end{subequations}
with $\mathbf{Q} = \operatorname{diag}(20, 20, 3)$, $\mathbf{R} = \operatorname{diag}(0.05, 0.08)$, $\mathbf{u}_{\min} = [0, -1.2]^\top$, and $\mathbf{u}_{\max} = [3.5, 1.2]^\top$ (linear velocity in m/s, angular rate in rad/s). Only the first control input is applied, and the problem is re-solved at each step via CVXPY solvers. Mean errors and control energies $\frac{1}{N}\sum_{k=0}^{N-1} \|\mathbf{u}_k\|_2^2$ ($N$ is control steps) are reported in Table~\ref{tab:tracking_energy}, where fallback indicates numerical failure.

\revised{\textbf{Covariance-regulated RBF methods demonstrate numerical survival and reliable tracking.}} Their mean errors are smaller and the control energies are comparable to the baseline. On the other hand, the vanilla and SIFt variants of the \texttt{Poly} dictionary recover after large transients (Fig.~\ref{fig:DDR_control_benchmark}(a)) but are less reliable. Hence, covariance regulation prevents the identified \ac{RBF} failure as previously revealed in Table~\ref{tab:ddr_online_modeling_results}. 

\revised{\textbf{Multi-step error is more informative than one-step error for closed-loop evaluation of predictive control.} For regulated \ac{RBF} methods, one-step learning performance and control performance rankings happen to agree (DZ-CS, SIFt, DZ-KS, CT). However, across dictionaries, this does not hold: vanilla, DZ-CS, DZ-KS have one-step errors near $1.96\times 10^{-3}$ but control errors 0.195, 36.3, and 9.7. Among all online variants, only SIFt demonstrates robustness across dictionaries, ranking first with \texttt{Poly} and second with \texttt{RBF} dictionary. This is likely due to its best multi-step error (Fig.~\ref{fig:sensitivity}(d)) and double-sided covariance bounds.}

\revised{\textbf{Low off-policy prediction error does not certify closed-loop control suitability.} 
The discrepancy between Tables~\ref{tab:ddr_online_modeling_results} and~\ref{tab:tracking_energy} 
arises because open-loop prediction and closed-loop control evaluate decoupled model properties. 
While one-step metrics (Table~\ref{tab:ddr_online_modeling_results}) and frozen-parameter rollouts (Fig.~\ref{fig:sensitivity}(d)) evaluate 
the observed dynamics along known trajectories, they fail to assess the 
counterfactual input responses evaluated by MPC. Because condensed MPC predictions rely on 
horizon-wise control sensitivities $\mathbf{C}\mathbf{A}_k^{j-1}\mathbf{B}_k$, errors in 
$\mathbf{A}_k$ and $\mathbf{B}_k$ can cancel out along recorded data, yielding small 
prediction residuals, yet produce erroneous gradients during optimization. The resulting 
feedback alters the closed-loop state--input distribution, triggering a distribution shift 
that is further amplified by the globally supported observables of the 
\texttt{Poly} dictionary. Empirically, the forward-speed lower constraint ($v_{\min}=0$) is active for 
77.4\% of \texttt{Poly} (DZ-CS) steps and 98.1\% of \texttt{Poly} (CT) steps, showing that their horizon 
models repeatedly drive the optimizer toward ineffective boundaries, halting the robot while the reference advances. This failure mode yields a problematic input-coupling 
matrix $\mathbf{B}$ with unphysical entries ($\frac{\partial \dot{p}_x}{\partial v} = -0.0365$, 
$\frac{\partial \dot{p}_y}{\partial v} = 0.0296$, $\frac{\partial \dot{\theta}}{\partial \omega} = 0.0477$), since $\partial \dot{p}_x/\partial v$ is positive for trajectory stays where $\cos\theta>0$ for a non-holonomic vehicle. 
}

\revised{\begin{remark}(Computation Time)
    Execution times were profiled on a MacBook M1 processor. State lifting and recursive operator updates required $0.15$--$0.28$~ms median ($0.22$--$0.41$~ms $95^{\text{th}}$ percentile), while the 50-horizon CVXPY solver required $14.8$--$18.5$~ms median ($21.2$--$32.4$~ms, $95$-th percentile). Total execution time remains well below the $50$~ms ($\Delta t = 0.05$~s) control deadline.
\end{remark}}

% \begin{table}[htbp]
% \centering
% \footnotesize
% \setlength{\tabcolsep}{10pt}
% % \setlength{\tabcolsep}{4pt}
% \renewcommand{\arraystretch}{0.8}
% \caption{Control Performance with Online Learning for a DDR}
% \label{tab:tracking_energy}
% \begin{tabular}{lcc}
% \toprule
% \textbf{Method} & \textbf{Mean Error} & \textbf{Control Energy} \\
% \midrule
% Offline RBF & $4.46 \times 10^{-2}$ & $3.53 \times 10^{0}$ \\
% Poly  & $2.43 \times 10^{-1}$ & $4.05 \times 10^{0}$ \\
% Poly (DZ-CS)& $2.05 \times 10^{1}$ & $7.60 \times 10^{0}$ \\
% Poly (DZ-KS) & $3.54 \times 10^{-1}$ & $4.20 \times 10^{0}$ \\
% Poly (CT) & $7.04 \times 10^{1}$ & $3.96 \times 10^{-1}$ \\
% RBF & $7.28 \times 10^{1}$ & $1.63 \times 10^{0}$ \\
% RBF (DZ-CS) & $3.85 \times 10^{-2}$ & $3.57 \times 10^{0}$ \\
% RBF (DZ-KS) & $3.93 \times 10^{-2}$ & $3.55 \times 10^{0}$ \\
% RBF (CT) & $4.26 \times 10^{-2}$ & $3.54 \times 10^{0}$ \\
% \bottomrule
% \end{tabular}
% \end{table}

\section{Conclusion}
\label{sec:conclusion}

This paper studied online recursive Koopman learning, revealing vulnerabilities to estimator freezing and covariance windup. Using a simulated ground vehicle and real-flight data from an FWMAV, we showed that vanilla recursive updates can improve adaptation under domain shifts but risk numerical failure. \revised{We introduced CR-RKL as a unified framework to systematically assess covariance-regulated mechanisms. We proposed a novel error dead-zone covariance skipping strategy, which conditionally freezes the covariance update while maintaining learning efficiency, outperforming variants that suspend the entire identification loop. We additionally adapted constant-trace normalization and directional forgetting strategies into Koopman formulation.} \revised{However, covariance regulation is not a universal remedy. We explicitly exposed a structural limitation: unbounded lifting functions paired with dead-zone gating or constant-trace normalization can lead to incorrect operator learning and MPC failure, making them strictly suited for bounded dictionaries.} Future work will validate CR-RKL on more diverse robotic systems.

\addtolength{\textheight}{-12cm}   % This command serves to balance the column lengths

\bibliography{reference}
\bibliographystyle{ieeetr}

\end{document}